\documentclass{ctic}

\usepackage{graphicx}

\usepackage{graphics} 
\usepackage{epsfig} 

\newtheorem{theorem}{Theorem}[section]

\newtheorem{lemma}[theorem]{Lemma}

\begin{document}

\title{Genus Computing for 3D Digital Objects: Algorithm and Implementation}

\author{Li Chen\\
\emph{University of the District of Columbia}\\
\emph{lchen@udc.edu}
}

\maketitle
\thispagestyle{empty}

\begin{abstract}

This paper deals with computing topological invariants such as connected components, 
boundary surface genus, and homology groups.   
 For each input data set, we have designed or implemented algorithms to 
 calculate connected components, boundary surfaces and their genus, and 
homology groups. Due to the fact that genus calculation dominates the entire task for 3D object in 3D
space, in this paper, we mainly discuss the calculation of the genus. The new algorithms designed in this paper
 will perform: 
(1) pathological cases detection and deletion, 
(2) raster space to point space (dual space) transformation, (3) the linear time algorithm for 
boundary point classification, and (4) genus calculation.

\end{abstract}


\section{Introduction}

Computing topological properties for a 3D object in 3D space is an important task in image processing.
The recent developments in medical imaging and 3D digital camera systems raise the problem of the direct 
treatment of digital 3D objects. In the past, 3D computer graphics and computational 
geometry have usually used triangulation to represent a
3D object.
  
Basically, the topological properties of an object in 3D contains connected components,
genus of its boundary surfaces, and other homologic and homotopic properties~\cite {Kaz03}. 
In 3D, this problem of obtaining fundamental groups is decidable but no practical 
algorithm has yet been found.
Therefore, homology groups have played the most significant role~\cite{Day98} ~\cite{Kaz04} . 
Research shows that a key factor of computing homology groups of 3D objects
is the genus of the boundary surface of the 3D object~\cite{Del95}.

Theoretical results show that there exist linear time algorithms for calculating genus and homology groups 
for 3D Objects in 3D space~\cite{Day98}. However, the implementation of these algorithms is not simple due to the complexity 
of real data samplings. Most of the algorithms require the triangulation of the input data since it is 
collected discretely. However, for most medical images, the data was sampled consecutively, meaning that
every voxel in 3D space will contain data. In such cases, researchers use the marching-cubes algorithm 
to obtain the triangulation since it is 
a linear time algorithm ~\cite{LC87}. However the spatial 
requirements for such a treatment will be at least doubled by
adding the surface-elements (sometimes called faces). 

The theoretical work of calculating genus based on simple
decomposition will turn into two different
procedures: (1) finding the boundary of a 3D object and then using polygon mapping, also called polygonal schema, 
(2) cell complex reductions
where a special data structure will be needed.

In this paper, we look at a set of points in 3D digital space, and
our purpose is to find homology groups of the data set. 
The direct algorithm without utilizing triangulation was proposed by Chen and Rong in 2008~\cite {CR08}
However, this algorithm 
is based on the strict definition of digital surfaces. Many real 3D sets may not satisfy the definition. 
In other words, a set of connected points may not be able to be put 
into such a process without considerable associated theoretical
and practical processes. 

In \cite {CR08}, we discuss the geometric and algebraic properties of manifolds 
in 3D digital spaces and the optimal algorithms for calculating these properties. 
We consider {\em digital manifolds} as defined in ~\cite {Che04}. 
More information related to digital geometry and topology 
can be found in ~\cite{KR} and ~\cite{KoR}.
We presented a theoretical optimal algorithm with
time complexity $O(n)$ to compute the genus and homology groups in 3D digital space, 
where $n$ is the size of the input data ~\cite {CR08}.  
  
The key in the algorithm in ~\cite{CR08} is to find the genus of the closed digital surfaces that
is the boundary of the 3D object.
However, the new algorithm is based on the strict definition of closed digital surfaces in \cite{Che04},
which means that there are many cases of real sampling of 3D objects that do not satisfy the definition
of digital surfaces. In this paper, we will also deal with extreme situations. We have designed an adding and deleting method to make the 3D object into manifolds. 

This paper provides a complete process that deals with simulated and real data in order to obtain
the topological invariants such as connected components, boundary surface genus, and homology groups.   
 
For every input data set, we have designed or implemented algorithms to 
calculate connected components, boundary surfaces and their genus, and 
homology groups. This is due to the fact that genus calculation dominates the entire task for 3D objects in 3D
space. In this paper we mainly discuss the calculation of the genus. The new algorithms designed in this paper
 will perform: 
(1) pathological cases detection and deletion, 
(2) raster space to point space (dual space) transformation, (3) the linear time algorithm for 
boundary point classification, and (4) genus calculation.   
 
This paper does not consider homology generators. For more details, please see \cite{Day98}\cite{Pel}.

\section {Concepts and Theoretical Results} 

In this section, we review some existing work related to this paper including the genus
of closed digital surfaces, homology groups of manifolds in 3D digital space, and
a theoretical linear algorithm of finding Homology Groups in 3D~\cite {CR08}.

\subsection{Genus of Closed Digital Surfaces}
 
Any continuous 3D object can be viewed as a collection of 3D voxels in digital or
cubical space.
Unless the sampling method is changed, any practical method of genus calculation must adapt to this fact. 
Medical imaging such as CT and MRI are such examples.     

Cubical space with direct adjacency, or (6,26)-connectivity in digital space\cite{Che04}, has the simplest 
topology in 3D digital spaces. It is also sufficient for the topological 
property extraction of 
digital objects in 3D. Two points are said to be adjacent in 
(6,26)-connectivity space if the 
Euclidean distance of these two points is 1, called direct adjacency.
 
Let $M$ be a closed (orientable) digital surface in the 3D grid space in direct adjacency. 
We know that there are exactly 6-types of digital surface
points. This was first discovered by Chen and Zhang in~\cite{CZ93}. Relation to different 
definitions of digital surfaces  can be found in 
\cite{CCZ99}.

\begin{figure}[h]
	\begin{center}

   \epsfxsize=2in 
   \epsfbox{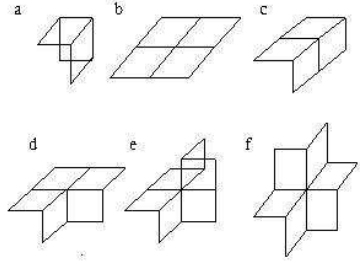}

	\end{center}
\caption{ Six types of digital surfaces points in 3D}
\end{figure}

Assume that $M_i$ ($M_3$, $M_4$, $M_5$, $M_6$) is the set of 
digital points with $i$ 
neighbors.   We have the following result for a simply connected 
$M$ ~\cite{CZ93}\cite{Che04}:

\begin{equation}
          |M_3| =8 + |M_5| + 2 |M_6| .      
\end{equation}
 
\noindent $M_4$ and $M_6$ has two different types, respectively.

The Gauss-Bonnet theorem states that if $M$ is a closed manifold, then

\begin{equation}
          \int_{M} K_{G} d A = 2 \pi \chi(M)    
\end{equation}

\noindent where $d A$ is an element of area and $K_{G}$ is the Gaussian curvature. 

Its discrete form is

\begin{equation}
          \Sigma_{\{p \mbox{ is a point in } M\}} K(p) = 2 \pi \cdot (2- 2 g)  
\end{equation}


\noindent where $g$ is the genus of $M$.

Assume that $K_i$ is the curvature of elements in $M_i$, $i=$ 3,4,5,6. We have

\begin{lemma}\label{l21}
   (a) $K_3  = \pi/2$,
  (b) $K_4= 0$,  for both types of digital surface points,
  (c) $K_5  = - \pi /2$, and
   (d) $K_6  = - \pi$,  for both types of digital surface points.
\end{lemma}

\noindent We obtained (see \cite{CR08}),

\begin{equation}
           g = 1+ (|M_5|+2 \cdot |M_6| -|M_3|)/8. 
\end{equation} 

The two examples show that the above formula  is correct \cite{CR08}. The first example shown in 
Fig. 2, is the easiest case.   

The second example came from the Alexander horned sphere. See Fig. 3.   

\begin{figure}[h]
	\begin{center}

   \epsfxsize=2in 
   \epsfbox{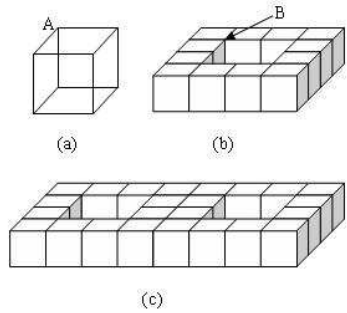}

	\end{center}
\caption{ Simple examples of closed surfaces with $g=0,1,2$}
\end{figure}

\begin{figure}[h]
	\begin{center}

   \epsfxsize=2.3in 
   \epsfbox{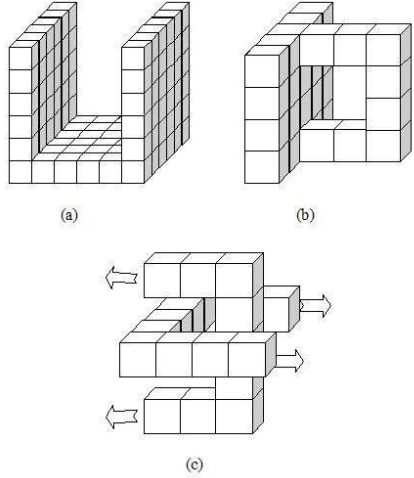}

	\end{center}
\caption{An example came from Alexander horned sphere in digital space}
\end{figure}

\subsection{Homology Groups of Manifolds in 3D Digital Space} 

For a $k$-manifold,
Homology group $H_{i}$, $i=0,...,k$ provides the information for the number of holes in each $i$-skeleton of the
manifold. When the genus of a closed surface is obtained, we can then
calculate the homology groups corresponding to its 3-dimensional
manifold in 3D. 

The following result follows from standard results in algebraic topology  \cite{Hat}.
It also appears in \cite{Day98}.  Let $b_i = \mbox{rank} H_i(M, Z)$ be the $i$th Betti number of $M$. 
The Euler characteristic of $M$ 
is defined by

\[ \chi (M) = \sum_{i\geq 0} (-1)^i b_i \]

If $M$ is a 3-dimensional manifold, $H_i(M)=0$ for all $i>3$ essentially because 
there are no $i$-dimensional holes.
Therefore, $\chi (M)  = b_o - b_1 + b_2 - b_3$.
 Furthermore, if $M$ is in
$R^3$, it must have nonempty boundary. This implies that $b_3 = 0$.

\begin{theorem}\label{Jordan2}
Let $M$ be a compact connected 3-manifold in $S^3$. Then
\begin{enumerate}
\item [(a)]  $H_0(M)\cong Z$.
 \item [(b)] $H_1(M) \cong Z^{\frac{1}{2} b_1(\partial M)}$, i.e. $H_1(M)$ is torsion-free with rank being half of rank $H_1(\partial M)$.
\item[(c)] $H_2(M) \cong Z^{n-1}$ where  $n$ is the number of components of $\partial M$.
\item[(d)] $H_3(M)=0$ unless $M=S^3$.
\end{enumerate}
\end{theorem}

A proof of above theorem is shown in \cite{CR08}.

\subsection{A Theoretical Linear Algorithm of Finding Homology Groups in 3D}

Based on the results we presented in the above subsections, we
now describe a linear algorithm for computing the homology group of 3D objects
in 3D digital space ~\cite{CR08}.

Assuming we only have a set of points in 3D. We can digitize this set into 3D digital spaces. 
There are two ways of doing so: (1) by treating each point as a cube-unit that is called the 
raster space, 
(2) by treating each point as a grid point, which is also called the point space.
These two are dual spaces.
Using the algorithm described in ~\cite{Che04}, we can determine whether the digitized set forms a 
3D manifold in 3D space in direct adjacency for connectivity. The algorithm is in linear time.

\noindent {\bf Algorithm 2.1} Let us assume that we have a connected $M$ that 
is a 3D digital manifold in 3D.

\begin{description}

 \item [Step 1.] Track the boundary of $M$, $\partial M$, which is a union of several closed surfaces. 
This algorithm only needs to scan though all the points in $M$ to see if 
the point is linked to  
a point outside of $M$. That point will be on boundary. 

\item [Step 2.]  Calculate the genus of each closed surface in $\partial M$ using the method 
described in Section 2. We just need to count the number of neighbors on a surface.  
and put them in $M_i$, using the formula (5) to obtain $g$.

\item [Step 3.] Using the Theorem \ref{Jordan2}, we can get $H_0$, $H_1$, $H_2$, and $H_3$. 
                $H_0$ is $Z$. For $H_1$, we need to get $b_1(\partial M)$ that is just 
				the summation of the genus in all connected components in  $\partial M$. (See \cite{Hat}
				and \cite{Day98}.) 
				$H_2$ is the number of components in $\partial M$. $H_3$ is trivial.
\end{description}

\begin{lemma}
           Algorithm 2.1 is a linear time algorithm. 
\end{lemma}

Therefore, we can use linear time algorithms to calculate $g$ and all homology
 groups for digital manifolds in 3D based on Lemma 2.2 and Lemma 4.1.

\begin{theorem}
           There is a linear time algorithm to calculate all homology
 groups for each type of manifold in 3D.  
\end{theorem}

\section {Algorithm Design and Implementation }

The algorithm described in Section 2 is a theoretical result. The implementation of the algorithm must consider
all possible cases in practical data collection. We first need to find the boundary and then decide if
the boundary is a 2D manifold. If the boundary data connecting voxel data
sets are not purely defined digital surfaces, we will have three options: (1) we need to modify the data 
to meet the requirement before genus calculation, (2) if the change of the original data set is too great, we
may need to stop the modification instead of outputting a result for reference, and (3) we make some limited changes,
and then produce a result.     

The difference between the theoretical results and practical data processing is that we may not always get the 
input data we expected. In our case, the boundary of a solid object should be
treated as a surface. However, practically, this might not always be the case. 
Some researchers also consider making real data sets ``well''-organized.  Siqueira {\it et al}
considered making a 26-connected data set well-composed~\cite{SLG05} 
\cite{SLTGG08}~\cite{BB08}. This means that two voxels will be connected
by a sequence of voxels where each pair of two adjacent cubes share a 2D-cell (face-unit).    
The concept of well-composed is mathematically equivalent to 
6-connected. An algorithm described in ~\cite{SLG05} 
\cite{SLTGG08} may generate new "none" well-composed cases, which are not good selections 
 for genus calculation.

Our new algorithm and implementation will perform: (1) pathological cases detection and deletion, 
(2) raster space to point space (dual space) transformation, (3) the linear time algorithm for 
boundary point classification, and (4) genus calculation.    

Some detailed considerations of recognition algorithms related to 3D manifolds can be 
found in ~\cite{BK08} where
Brimkov and Klette made extensive investigations in boundary tracking.  The discussions 
of 3D objects in raster space can be found in ~\cite{Lat}.

\subsection{Input Data Sets}

This subsection will discuss the input data formats. We will focus on cubical data, 
for instance MRI and CT data.
In cubical data samples, we assume that the sampling is contiguous, where each sample point is 
normally followed by 
another sample point in its neighborhood. It is important to know this because a random sampling can cause 
the problem of uncertainty.  
In this case, we usually
cannot calculate the genus without making an assumption. For instance, we will not be able to know 
where a hole is. 
In order to get simplicial decomposition (usually triangulation), we usually need to use 
Voronoi or Delaunay decomposition
with boundary information. That means the boundary must be assumed. 

A new technology is called persistent homology analysis that tells us how to find the best estimation
 for the location 
of holes, usually by multiscaling (the upscaling and downscaling methods). However, this method is not a 
precise analysis ~\cite{Car}~\cite{ZC}.  
  
Even though, our method can be modified to be used in persistent analysis, this paper mainly deals with
the method of precise genus and homology group calculation. 

In summary, our assumption is that the digital object consists of cubical points (digital points, raster points). 
Each point is a cube, which is the smallest 3D object. The edge and point are defined with regards to the cube and
an object may contain
several connected components using a cube-linking path. Our purpose again is to calculate the 
topological properties
of the object, or of each component, essentially. 

\subsection{Searching connected components of a cubical data set}

Connected component search is an old task that can be done by using Tarjan's Breadth-first-search. 
Pavlidis was one of  
the first people to realize and use this algorithm in image processing. This problem is also known as the 
labeling problem. The complexity of the algorithm is $O(n)$ ~\cite{Pav}. 

The problem is what connectivity is based off of. In 3D, we usually have 6-, 18-, 26- connectivity. Since real data has
noise, we have to consider all of those connectivities. So we must use 26-connectivety to get the connected components.

Therefore, the connected component of the real processing is not a strictly 6-connected component. The topological 
theorem generated previously in ~\cite{CR08} is no longer suitable. So we need to transform 
a 26-connected component into a 6-connected component. This should be done by a meaningful adding or deleting process since
optimization on the minimum number of changes could be an NP-hard problem. 

{\bf Problem of minimum modifications:} Given a set of points in 3D digital space, if this set is not a manifold, assume that the points are connected in 
a connectivity defined using adding or deleting processes to make the set
a 3D manifold. The question becomes: is there a polynomial algorithm that makes  the solution have minimum modifications where adding or deleting 
 a data point will be counted as one modification?

A similar problem was considered in ~\cite{SLG05} in which a decision problem of adding was proposed. 

This problem can be extended to a general $k$-manifold in $n$-D space. Even though we have the 6-connected 
component, there may still be cases that 
contain the pathological situation, which needs special treatment. We will discuss this issue in 
the next subsection.

\subsection{Pathological Cases Detection and Deletion}

In this paper, we only deal with the Jordan manifolds, 
meaning that a closed $(n-1)$-manifold will separate the $n$-manifold into two or more components. 
For such a case, only direct adjacency will be allowed since indirect adjacency will not generate
 Jordan cases. 

That is to say, if the set contains indirect adjacent voxels, 
we need to design an algorithm to detect the situation and delete some voxels in order to preserve  the homology groups.

It is known that there are only two such cases in cubical or digital space \cite{Che04}: two 
voxels (3-cells) share 
a 0-cell or a 1-cell. Therefore, we want to modify the voxel set to only contain voxels where two of these cases 
do not appear. Two voxels share exactly a 2-cell, or there is a local path (in the neighborhood) 
of voxels where two adjacent
voxels share a 2-cell ~\cite{Che04}. A special case was found in \cite{SLTGG08} that is the complement case of
the case in which two voxels share a 0-cell (see Fig 4. (a)). The case may create a tunnel or could be filled. We will simplify it by 
adding a voxel in a $2\times 2\times 2$ cube. Such a case in point space is similar to the case (a) in Fig 4 since
the boundaries of these cases are the same.  
 
The problem is that many real data sets do not satisfy the above restrictions (also called well composed
image). The detection is easy but deleting
certain points (the minimum points deletion) to preserve the homology is a bigger issue.

The following rules (observations) are reasonable: In a neighborhood $N_{27}(p)$ that contains 8 cubes and
27 grid points,

 a) if a voxel only shares a 0-cell with a voxel. This voxel can be deleted.  

 b) if a voxel only shares a 1-cell with a voxel. This voxel can be deleted.  

 c) if a boundary voxel $v$ shares a 0,1-cell with a voxel, assume $v$ also shares a 2-cell with a voxel $u$, 
    $u$ must share a 0,1-cell with a voxel that is not in the object $M$.  $u$ is on the boundary. 
    Deleting $v$ will not change the topological properties.
    
 d) if in a $2\times 2\times 2$ cube, there are 6 boundary voxels and its complement (two zero-valued voxels) 
    is the case (a) in Fig (4). Add a voxel to this $2\times 2\times 2$ cube such that the new voxel
    shares as many 2-cells in the set as possible. This means that we want the adding voxel to be inside of
    the object as much as we can.

\begin{figure}[h]
	\begin{center}

   \epsfxsize=3in 
   \epsfbox{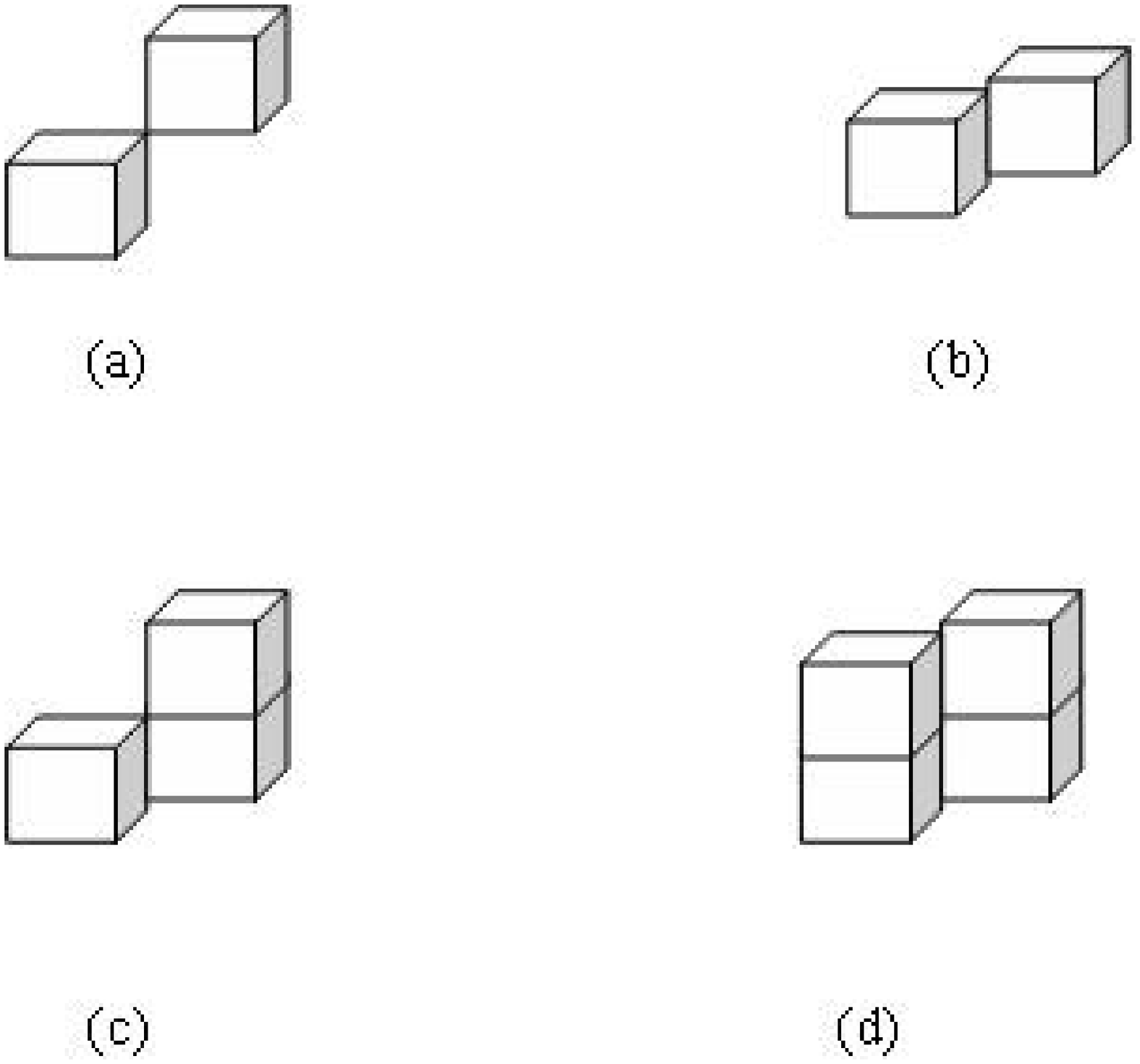 }

	\end{center}
\caption{ Pathological Cases}
\end{figure}

In this paper, we implement or modify the above rules to fit the 
theoretical definition of the digital surfaces.  We also design an algorithm based on these 
rules to detect and delete some data points while preserving 
the topology. This is essential to calculating the genus correctly.  However, when the object becomes more
complex, pathological situations may still exist.

\begin{figure}[h]
	\begin{center}

   \epsfxsize=4in 
   \epsfbox{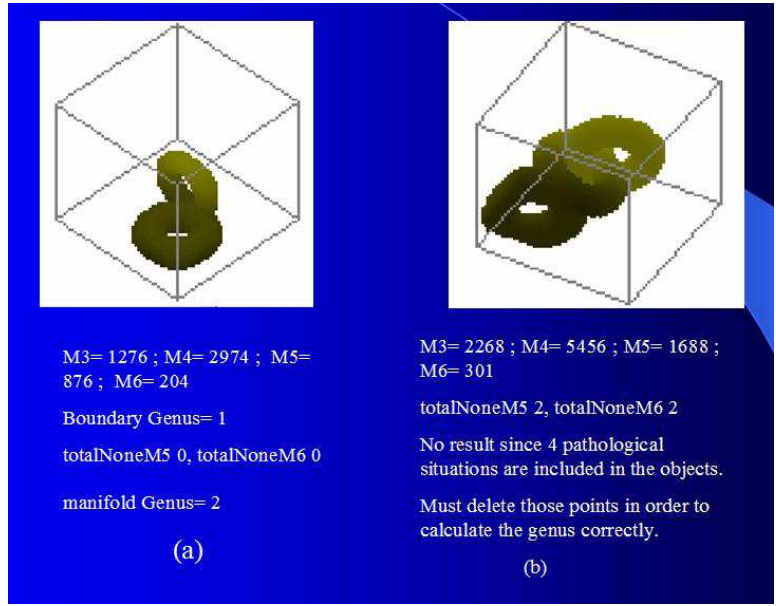}

	\end{center}
\caption{ Without Pathological Case Process}
\end{figure} 

The mathematical foundation of this above process that eliminates pathological cases is still under investigation. 

{\bf Mathematical foundation of Modifying a 3D object to be a 3D manifold:} Given a set of points in 3D 
digital space,
how would we modify the data set into a manifold without losing or changing the topology (in mathematics)?

\subsection{Boundary Search}

  In general, a point is on the boundary if and only if
it is adjacent to one point in the object and one point not in the object (in 26-connectivity). 
A simple algorithm that goes through each point and tests the neighborhood will determine whether a point is on 
the boundary or not. This is a linear time and $O(log(n))$ space algorithm. 
 
The only thing special is that we use 26-connectivity to determine the boundary points. This is to take all
possible boundary points into consideration in the next step.

\subsection{Determination of the Configuration of Boundary Points} 
 
When all boundary points are found, we need to find their classifications. In other words, we need to determine whether
a special point is in $M_3$, $M_4$, $M_5$, or $M_6$. Here is the problem: if we only have one voxel, is it 
a point (0-cell) or a 3D object (3-cell)? In this paper, we treat it as a 3-cell. 

The input data is in raster space, but the boundary surface will be in point space. We must first make the 
translation. Then, for each point on the surface, we count how many neighbors exist in order to determine its configuration 
category. After that, we use formula (4) to get the genus. 

If we still need to find homology groups, we can just use the simple calculations based on Theorem 2.3 to 
obtain them. Using the program, we get the genus$=6$ for a modified real image (Fig. 7).

\begin{figure}[h]
	\begin{center}

   \epsfxsize=3in 
   \epsfbox{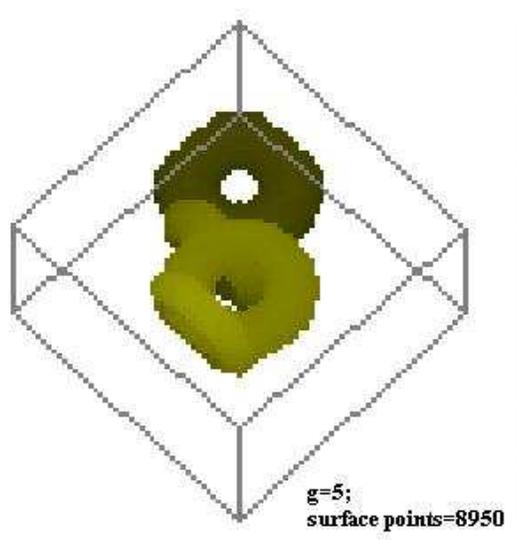}

	\end{center}
\caption{ After Pathological Case Process}
\end{figure}  
 
 \begin{figure}[h]
	\begin{center}

   \epsfxsize=3in 
   \epsfbox{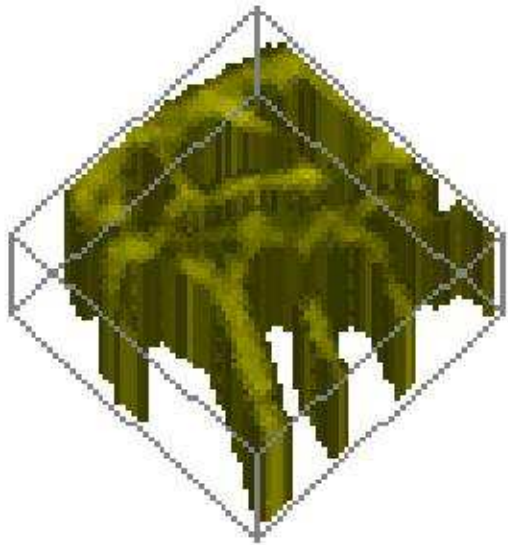}

	\end{center}
\caption{ A Modified Real Image}
\end{figure}








\end{document}